\documentclass{SCIS2024}

\usepackage{cite}

\usepackage{multirow}
\usepackage{colortbl} 
\usepackage{color}
\usepackage{xcolor}

\usepackage{makecell}
\usepackage{multirow}
\usepackage[percent]{overpic}

\def\eg{\emph{e.g.}}

\def\etal{{\em et al.~}}
\definecolor{mygray1}{gray}{.75}

\begin{document}
\ArticleType{RESEARCH PAPER}
\Year{2024}
\Month{}
\Vol{}
\No{}
\DOI{}
\ArtNo{}
\ReceiveDate{}
\ReviseDate{}
\AcceptDate{}
\OnlineDate{}

\title{COMPrompter: reconceptualized segment anything model with multiprompt network for camouflaged object detection}

\author[1]{Xiaoqin Zhang}{}
\author[1]{Zhenni Yu}{}
\author[1]{Li Zhao}{}
\author[2]{\\Deng-Ping Fan}{}
\author[3]{Guobao Xiao}{x-gb@163.com}
 
\AuthorMark{Zhang}

\AuthorCitation{Zhang, Yu, Zhao, Fan, Xiao}

\address[1]{Zhejiang Province Key Laboratory of Intelligent Informatics for Safety and Emergency,
Wenzhou University, Wenzhou 325035, China}
\address[2]{Nankai International Advanced Research Institute (SHENZHEN FUTIAN), 518045, \text{\&} CS, Nankai University}
\address[3]{School of Computer Science and Technology, Tongji University, Shanghai 201804, China}

\abstract{
We rethink the segment anything model (SAM) and propose a novel multiprompt network called COMPrompter for camouflaged object detection (COD). SAM has zero-shot generalization ability beyond other models and can provide an ideal framework for COD. Our network aims to enhance the single prompt strategy in SAM to a multiprompt strategy. To achieve this, we propose an edge gradient extraction module, which generates a mask containing gradient information regarding the boundaries of camouflaged objects. This gradient mask is then used as a novel boundary prompt, enhancing the segmentation process. Thereafter, we design a box-boundary mutual guidance module, which fosters more precise and comprehensive feature extraction via mutual guidance between a boundary prompt and a box prompt. This collaboration enhances the model’s ability to accurately detect camouflaged objects. Moreover, we employ the discrete wavelet transform to extract high-frequency features from image embeddings. The high-frequency features serve as a supplementary component to the multiprompt system. Finally, our COMPrompter guides the network to achieve enhanced segmentation results, thereby advancing the development of SAM in terms of COD. Experimental results across COD benchmarks demonstrate that COMPrompter achieves a cutting-edge performance, surpassing the current leading model by an average positive metric of 2.2\% in COD10K. In the specific application of COD, the experimental results in polyp segmentation show that our model is superior to top-tier methods as well. The code will be made available at https://github.com/guobaoxiao/COMPrompter.}

\keywords{Segment Anything Model, Camouflaged object detection, Boundary, Prompt}

\maketitle

\section{Introduction}
\label{sec:introduction}
Camouflaged object detection (COD)~\cite{fan2023advances} has been extensively studied as a subset of image segmentation tasks. It finds various applications in medical image segmentation~\cite{fan2020pranet}, nature conservation and wildlife research~\cite{lidbetter2020search}, and search and rescue missions.
\begin{figure}[t]
\centering
\begin{overpic}[width=0.7\linewidth]{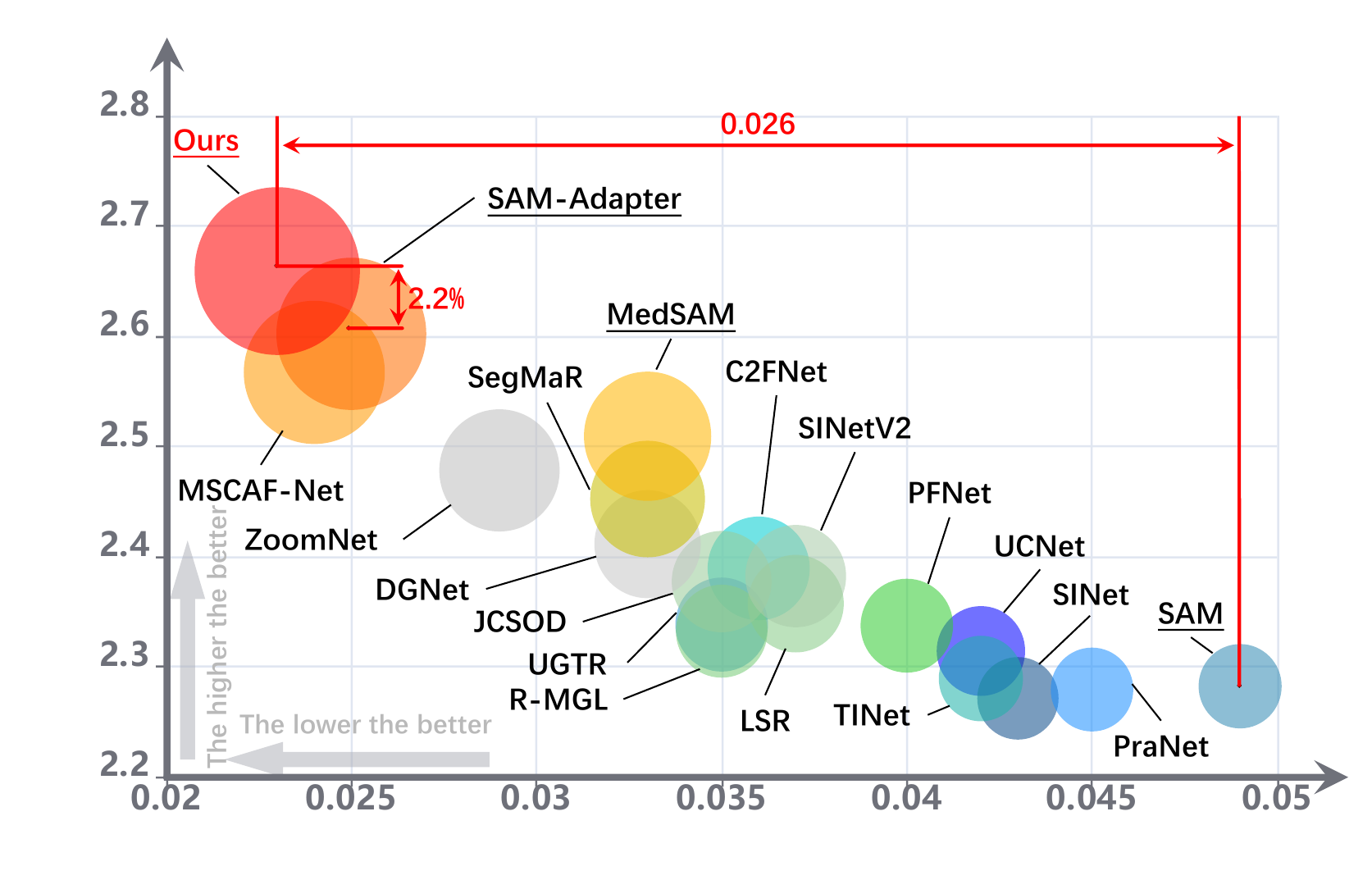} 
    \put(3.5,21){\small \rotatebox{90}{\textbf{Performance (Score)}}} 
    \put(34,1.5){\small  {\textbf{Mean Absolute Error (MAE)}}}
\end{overpic}
\vspace{-5pt}
\caption{\small{\textbf{Scatter plot representing the performance of competitors and our model on COD10K-Test.} $F_\beta^\omega$, $S_\alpha$, and $E_\phi$ are positive-oriented, while $M$ is negative-oriented. The order of magnitude of $M$ and the other indices are different. For a more effective comparison, we take $M$ as the X-axis, and the sum of the other three indicators as the Y-axis. The underline represents the segment anything model (SAM)-based method. (Score = $F_\beta^\omega$ + $S_\alpha$ + $E_\phi$).}}
\label{fig:begin}
\end{figure}

In the field of camouflaged object detection (COD), diverse methods are focused on essential information sources, including context (e.g., MSCAF-Net~\cite{liu2023mscaf}, C2FNet~\cite{sun2021context}), edge (e.g., JCSOD~\cite{li2021uncertainty}, R-MGL~\cite{zhai2021mutual}, TINet~\cite{zhu2021inferring}), and gradient (e.g., DGNet~\cite{ji2023deep}). Other methods employ a range of effective strategies such as amplification (e.g., ZoomNet~\cite{pang2022zoom}, ZoomNeXt~\cite{10568374}), humans attention (e.g., SegMaR~\cite{jia2022segment}), predation (e.g., SINetV2~\cite{fan2021concealed}, LSR~\cite{lv2021simultaneously}, PFNet~\cite{mei2021camouflaged}, PraNet~\cite{fan2020pranet}, SINet~\cite{9156837}), and uncertainty (e.g., UGTR~\cite{yang2021uncertainty}, UCNet~\cite{zhang2020uc}). In the recent studies, SAM-Adapter~\cite{chen2023sam} and MedSAM~\cite{ma2023segment} leveraged SAM~\cite{Kirillov_2023_ICCV} to perform COD. SAM excels in segmentation across various scenarios, including camouflage, because of its robust zero-shot generalization ability. SAM accomplishes COD with preliminary segmentation results at minimal computational cost. This capability enables researchers to develop more customized approaches for the unique characteristics of camouflaged targets. However, despite these strides, the existing methods often ignore the constraints associated with a single prompt. More critically, these methods fail to explore alternative prompt types other than those provided by SAM. In the context of COD, a noticeable disparity persists between SAM-based methods and the current state-of-the-art (SOTA) methods.

This study proposes COMPrompter, a multiprompt network for COD. COMPrompter leverages the strengths of SAM and expands  the utility thereof to the COD domain. Deviating from the direct use of SAM, COMPrompter introduces a multiprompt strategy. This strategy integrates both the original box prompt of SAM and the boundary prompt. In the boundary prompt, we particularly emphasize edges and gradients because of their extensive exploration. Identification of edges is more straightforward than recognition of an entire camouflage target, and gradients offer a fresh perspective on segmentation. Still, both approaches suffer their challenges: designing an edge module considerably increases computational overhead, and the targeted nature of internal gradients for various objects also poses limitations. Similar challenges exist in the specific application domain of COD, such as polyp segmentation.

Interestingly, we do not perform edge prediction in response to these challenges. Instead, we integrate edge information with gradients into the network via prompts, resulting in increased accuracy. Our proposed boundary prompt prevents the abovementioned issues. In addition, in conjunction with a box prompt, it provides a more accurate prior for COD. Specifically, the boundary prompt is derived from our proposed edge gradient extraction module (EGEM). EGEM employs dilation and canny operations on ground truth (GT) and image, respectively. Thereafter, EGEM obtains edge masks containing gradient instead of the entire camouflage target. Acquiring the gradient-enhanced boundary operation is straightforward yet innovative because prior research has not emphasized the gradient at the edge. To effectively guide segmentation using both the box and boundary prompts, we introduce the box-boundary mutual guidance module (BBMG). BBMG strengthens the connection between the dense box embedding and dense boundary embedding via adapted pointwise convolution of depthwise separable convolution~\cite{Laurent2014Rigid} and residual connection. In addition, inspired by He \etal~\cite{he2023camouflaged}, we incorporate the discrete wavelet transform (DWT) to obtain high-frequency signals. These signals represent details or rapidly changing parts of the image.

\begin{figure*}[t!]
\centerline{\includegraphics[width=\textwidth]{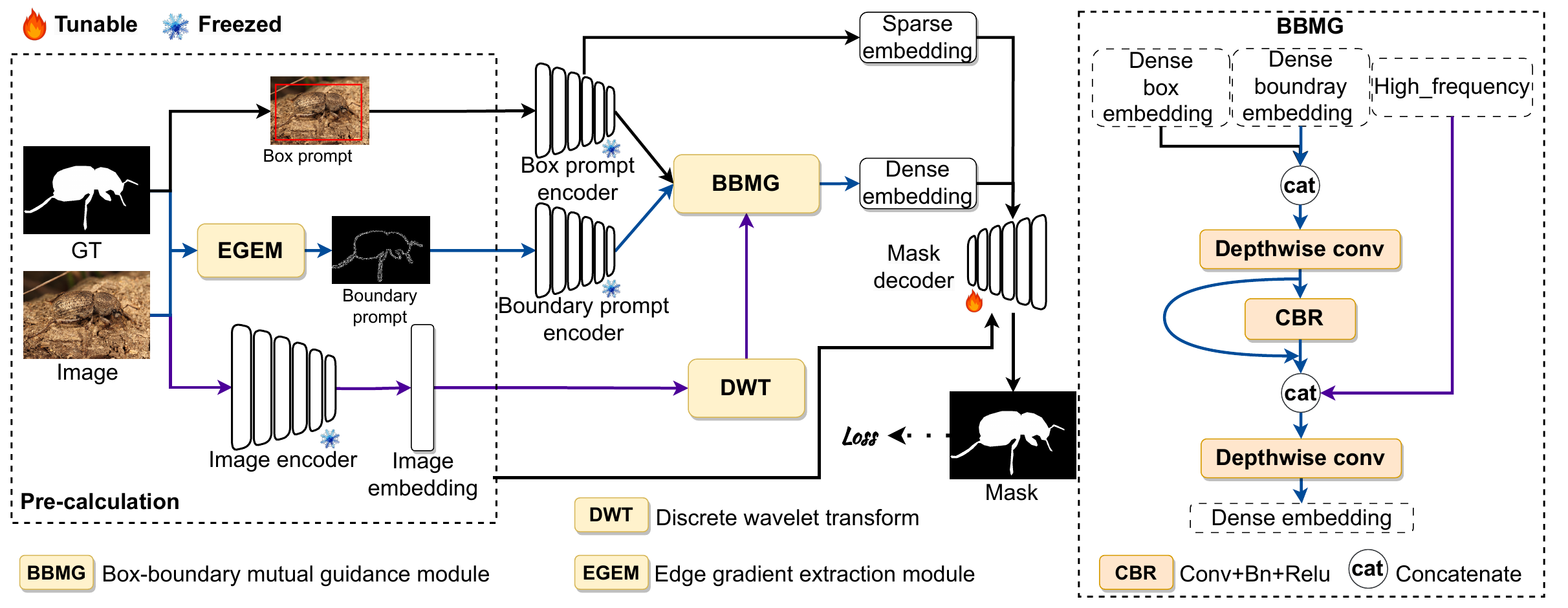}}
\vspace{-10pt}
\caption{\small{\textbf{Pipeline of our COMPrompter framework (left) and details of the box-boundary mutual guidance module (BBMG) (right).} Regarding the modules in SAM, the parameters in the module with a snowflake are fixed, while whose in the module with a spark can be optimized via training. Purple arrows represent image processing, while blue ones represent the processing of boundary prompts. The dashed arrow represents loss calculation. To decrease the amount of computation, we have calculated in advance the part in the left dashed box.}}
\label{fig:Mprompter}
\end{figure*}

Finally, through a judicious combination of EGEM, BBMG, and introduced DWT, we present COMPrompter for COD, a SAM variant tailored for COD. In COMPrompter, we adjust the SAM structure to accommodate our multiprompt strategy. Explicitly, we integrate a prompt encoder into SAM to address the proposed boundary prompt. Experimental results on four benchmark datasets substantiate the superiority of our method to all other SOTA methods, as shown in Fig.~\ref{fig:begin}. Fig.~\ref{fig:Mprompter} shows the complete network structure. 

Our contributions are summarized as follows:
\begin{itemize}
\item We propose a multiprompt network called COMPrompter for COD, a structural variant of SAM. Precisely, we propose a multiprompt strategy, in which the original box prompt of SAM and the newly designed boundary prompt are used as the user prompt of COMPrompter.
\item We propose two efficient designs in COMPrompter: EGEM and BBMG. EGEM obtains the gradient mask of the boundary from the image and GT. A box prompt and boundary prompt guide and complement each other via BBMG for accurate prompts.
\item We confirm the performance of COMPrompter on COD benchmark datasets. COMPrompter is observed to outperform the existing SOTA methods. We also conduct extensive experiments in polyp segmentation, and conclude that COMPrompter achieves a the cutting-edge performance in this domain.
\end{itemize}

\vspace{-10pt}
\section{Related work}\label{sec:related}
\vspace{-5pt}
\begin{figure}[t!]
\centerline{\includegraphics[width=.65\textwidth]{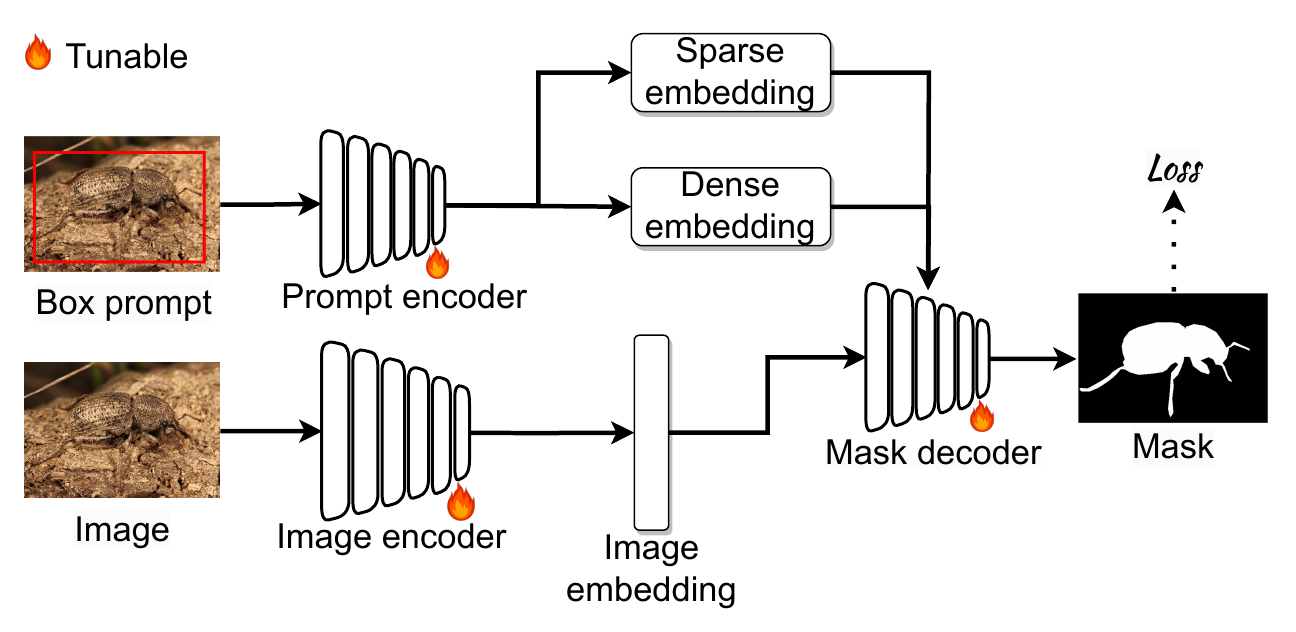}}
\vspace{-10pt}
\caption{\small{\textbf{The overview of SAM~\cite{Kirillov_2023_ICCV}} with box prompt.}}
\label{fig:SAM}
\end{figure}
\subsection{Segment anything model}
\vspace{-5pt}
As a new foundation model, SAM uses a massive dataset for training and has a remarkable zero-shot ability.
As shown in Fig.~\ref{fig:SAM}, SAM comprises three modules: image encoder, prompt encoder and mask decoder. The image encoder is large-scale pretrained via masked auto-encoder modeling and has satisfactory feature extraction ability. The prompt encoder encodes the user prompt to obtain sparse embedding and dense embedding. Important for the mask decoder is the design of the self-attention and cross-attention mechanisms. 
However, SAM often splits objects with the same semantic information into multiple masks~\cite{ji2023sam} because SAM lacks the guidance of semantic information.
In a specific field, SAM is unable to accurately segment fine structures because of a lack of professional knowledge or lack of a strong prior~\cite{ma2023segment}. Ma \etal~\cite{ma2023segment} employed a box prompt as a manual prompt, fine-tuned mask decoder, and achieved a satisfactory improvement in segmentation of fine structures. Chen \etal~\cite{chen2023sam} used an adapter to adapt SAM to COD and achieved satisfactory, if not remarkable, results.
On account of the inherent complexity of COD tasks, there remains significant room for improving the application of SAM in the COD domain. We propose a multiprompt strategy that leverages both a box prompt and a novel boundary prompt. The boundary prompt, a concept we propose, captures critical edge gradient information regarding the target object. The multiprompt strategy enhances the precision of prompts, thereby adapting to the difficulty of COD segmentation.

\subsection{Camouflaged object detection}
\vspace{-5pt}
Object detection~\cite{10149422}\cite{peng2022small} is a crucial task in computer vision. It aims to identify specific objects present in images and determine their locations. Its application in videos involves object tracking~\cite{jiang2024box}. The subtask of detection of  camouflaged targets is called COD~\cite{10007893}~\cite{10234216}. Simultaneous detection of camouflaged objects with similar properties across a set of images is called collaborative COD~\cite{10298243}. Several COD approaches integrate base models. Huang \etal~\cite{huang2023feature} focused on locality modeling and feature aggregation to mitigate the limitations of transformers. Luo \etal~\cite{chen2023diffusion} employed a diffusion model to generate salient objects in camouflaged scenes for training on multipattern images. Some COD methods give attention to the visual features (color, texture, brightness, etc.) of an object. Compared with these approaches, the strategy of giving attention to boundaries in COD is being increasingly accepted on a wide basis.
Zhu \etal~\cite{zhu2022can} designed a boundary guider module to accurately highlight the boundaries of hidden objects. Zhai \etal~\cite{zhai2021mutual} designed specified modules to enhance the visualization of edges. Ji \etal~\cite{ji2022fast} obtained an initial edge prior via selective edge aggregation. Sun \etal~\cite{sunboundary} employed excavation and integration of boundary-related edge semantics to increase the efficacy of COD. Lyu \etal~\cite{10183371} decoupled uncertainty reasoning and boundary estimation into two branches: uncertainty and boundary-guided features. These branches were then effectively aggregated to provide accurate segmentation information. Sun \etal~\cite{sun2023edge} proposed EAMNet, comprising an edge detection branch and a segmentation branch. The edge detection branch provided enhanced foreground representations, thereby acilitating the edge detection process. Dong \etal~\cite{dong2023unified}, grounded in the unified query-based paradigm, proposed UQFormer, which employed queries to derive boundary cues.
Meanwhile, object gradient generation was also applied in COD as an auxiliary task. Ji \etal~\cite{ji2023deep} mined texture information by learning object-level gradients. The application of object-level gradients for obtaining texture information is more deterministic than boundary modeling. It eliminates potential noise due to modeling.

However, a single boundary provides limited information. When providing the gradient of an entire target, the features learned by the network are messy because of the various types of camouflaged targets. Therefore, this study proposes a novel boundary mask with gradient information of the object-background junction. Compared with gradient information of an entire object, gradient information of object edges is easier to learn.

\section{Methodology}\label{sec:method}
\vspace{-5pt}
In this section, we present the details of multiprompt network (COMPrompter). First, we describe the overall architecture of COMPrompter, as shown in Fig.~\ref{fig:Mprompter}. Thereafter, we explain the core of this study, i.e., boundary prompt. Last, we discuss the necessity of introducing DWT.

\subsection{Overall architecture}
\vspace{-5pt}
The feature extraction part of COMPrompter can be categorized into two parts: the prompt encoder part and the image encoder part. Inspired by~\cite{ma2023segment}, we freeze the image encoder and prompt encoder and fine-tune the mask decoder. In addition, the image encoder is precomputed and stored as an npz file. The results of the image encoder can be directly read during real training. In this manner, the calculation amount and training threshold of the SAM large model are considerably decreased. We design the boundary prompt branch, which contains EGEM, BBMG and a parameter frozen prompt encoder. Among these, EGEM is designed to be computed in advance. The resulting image embedding of the image encoder is saved as an npz file, so that it can be directly read later. The boundary prompt combined with the original box prompt is used as the user prompt of COMPrompter. During inference, we use the GT to generate boundaries and boxes as user prompts to simulate scenarios of user interaction. We apply DWT to the image embedding to complement the frequency details of boundary features.

\subsection{Boundary prompt}
\vspace{-5pt}
We design a new boundary prompt branch with main modules comprising EGEM and BBMG. We first obtain the boundary mask with gradient by image, GT, edge detection and other operations. EGEM takes the boundary mask with gradient as input to the second prompt encoder to obtain the boundary embedding. Boundary prompt compensates for the original box prompt’s inability to provide precise boundary information. Specifically, the original box embedding and boundary embedding are fused by BBMG to achieve mutual guidance. Finally a key embedding is obtained as the input of the mask decoder.
\begin{figure}[t]
\centerline{\includegraphics[width=0.68\textwidth]{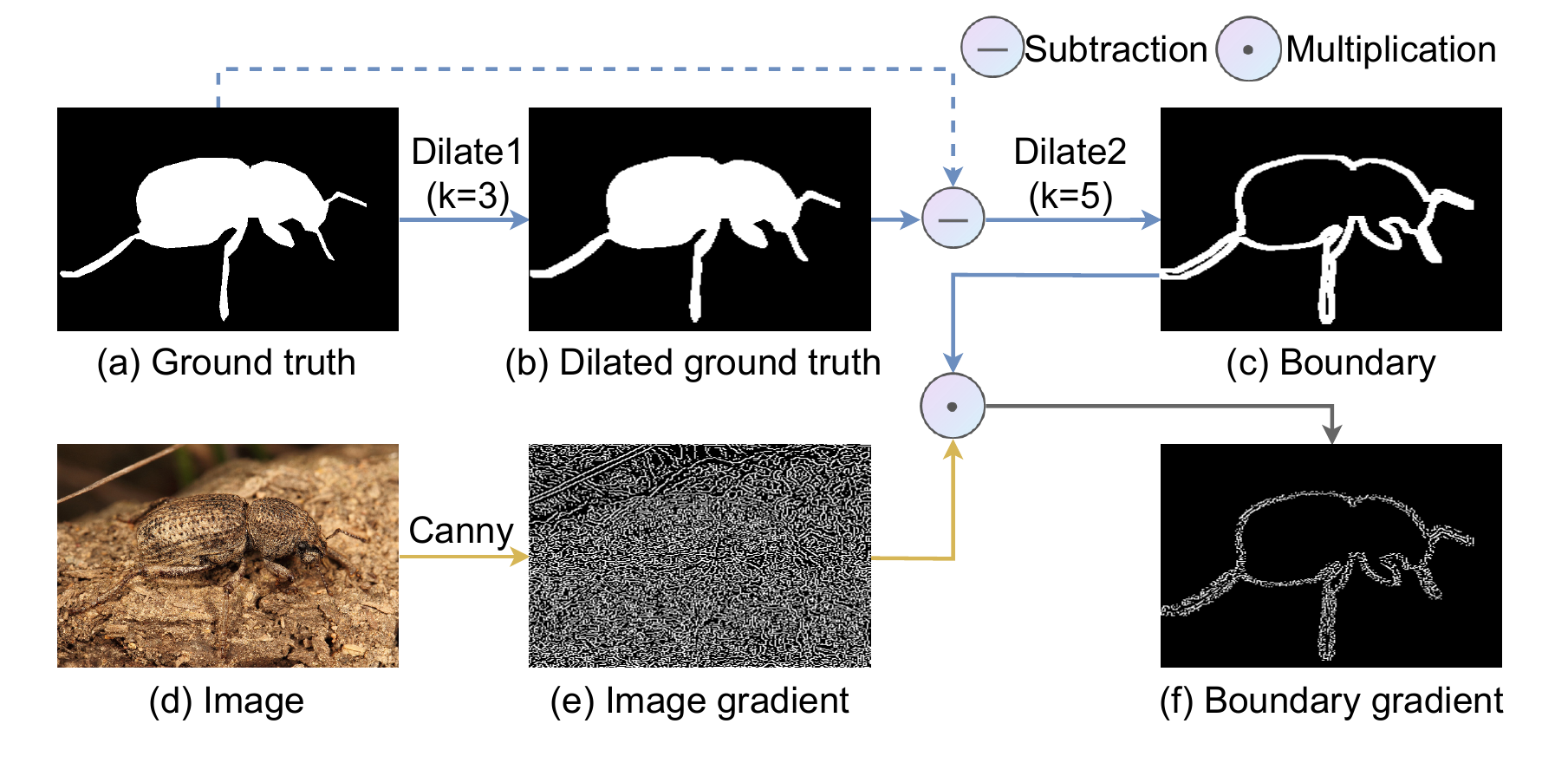}}
\vspace{-10pt}
\caption{\small{\textbf{EGEM details.} The top half of the figure represents the process of object edge extraction. The bottom half represents the process of extracting the gradient of the whole image. Finally the two images are multiplied to obtain the edge map containing the gradient.}}
\label{fig:EGEM}
\end{figure}

\subsubsection{Edge gradient extraction module}
\vspace{-5pt}
The boundary mask indicates the masks containing the gradient between the object and the background. Because the target is disguised, direct identification of the target is in COD. However, the difficulty of identifying the boundary is relatively easy, 
so boundary detection is added to COD as an auxiliary task~\cite{zhai2021mutual}. Gradient information is also widely used as another important information source. Ji \etal~\cite{ji2023deep} mined texture information by learning object-level gradients. However, compared with the gradient of the overall object, the gradient of the boundary undergoes a more rapid change. In addition, these gradients are more representative. Hence, we propose a boundary mask with a gradient.
As shown in Fig.~\ref{fig:EGEM}, EGEM employs operations such as dilate and canny to extract boundary masks with gradients from images and GT. The mask is the input of the prompt encoder.
Specifically, we first perform the dilation operation of GT with $kernel=3$, and the image obtained is subtracted  from the GT. In contrast to Premachandran \etal ~\cite{premachandran2017pascal}, wherein a pixel-wide boundary was employed, we use a dilation operation with a $3 \times 3$ kernel to obtain a thicker boundary. This choice is motivated by the necessity for not only the boundary but also its gradient. An excessively narrow boundary might lack sufficient information. Conversely, an excessively wide boundary might extend beyond the desired range and introduce noise. At this point, we obtain the edge image of the camouflaged target in the image. For learning more boundary information, this study performs the dilation operation of GT with $kernel=5$ on the obtained edge image to expand the edge of the image. In the prediction, the dilation operation of GT with $kernel=5$ also models the case wherein the user prompt may be inaccurate. The resulting boundaries are broader compared to the actual ones. This decreases the difficulty of obtaining boundaries at inference. The expanded edge image is multiplied with the canny image to obtain the final edge image with gradient information, the boundary mask.
The boundary mask with gradient $BG$ is generated as follows:
\begin{equation} \label{equ:EGEM}
BG = \gamma \big(\mu (GT) - GT \big) * C(I),
\end{equation}
where $\gamma (\cdot)$ and $\mu (\cdot)$ denote dilation operations with $kernel=5$ and $kernel=3$, respectively, $*$ multiplication, $C (\cdot)$ canny operation, and $I$ original image corresponding to $GT$.

In addition, we adopt the strategy of precalculation of EGEM like that in the image encoderbecause of EGEM’s independence from subsequent modules, decreasing the amount of calculation during training. Thereafter, we store the calculated images and features in npz files rather than common image formats. 

\subsubsection{Box-boundary mutual guidance}
\vspace{-5pt}
We propose BBMG with the intention of making a box prompt and boundary prompt guide each other and fuse each other. The box prompt is a sparse prompt, while boundary prompt is a dense prompt. The box prompt indicates the location of the target in the form of four points. The boundary prompt employs a mask to segment the boundary of the target to compensate for the lack of boundary details in the box prompt. Dense box embedding and dense boundary embedding are the results of the box and boundary prompts after the prompt encoder, respectively. As shown in the right-hand side of Fig.~\ref{fig:Mprompter}, we use the dense box embedding and dense boundary embedding to perform join operation on the channel dimension. Thereafter, we apply a residual operation and pass through basic units of convolution, batch normalization, and ReLU (CBR module), denoted as $CBR (\cdot)$. The optimized box-boundary embedding ($OBB$) is generated as follows:
\begin{equation}
\label{equ:BBMG}
EM = DC\big(cat(E_{box},E_{boundary})\big),
\end{equation}
\begin{equation}
\label{equ:BBMG2}
{OBB} = cat\big(CBR(EM), EM),
\end{equation}
where $E_{box}$ represents dense box embedding and $ E_{boundary}$ dense boundary embedding. $cat (\cdot)$ denotes the join operation on channel dimension. $DC (\cdot)$ represents an adapted pointwise convolution of depthwise separable convolutions~\cite{Laurent2014Rigid}.

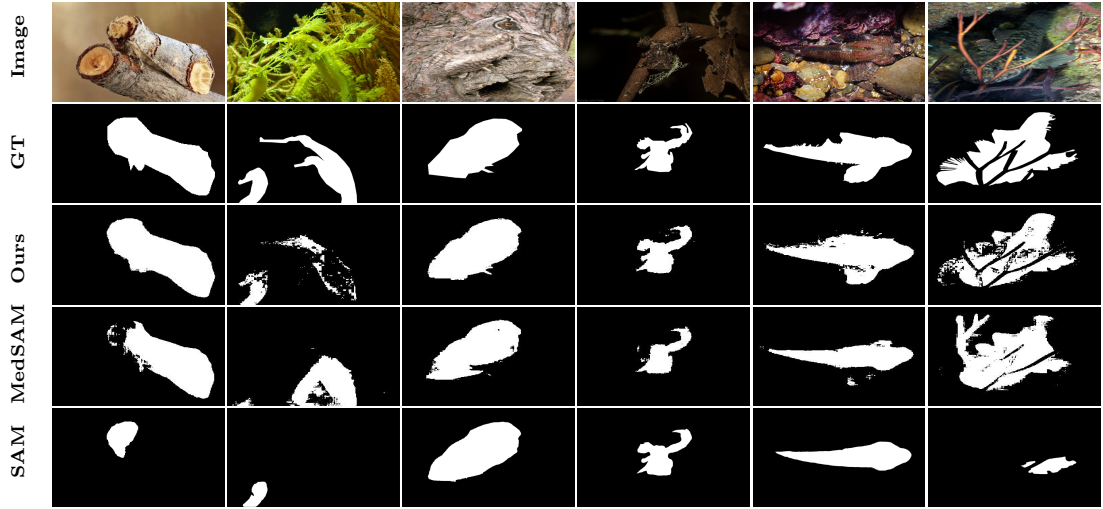
\begin{figure*}[t!]
\centering
\begin{overpic}[width=0.95\linewidth]{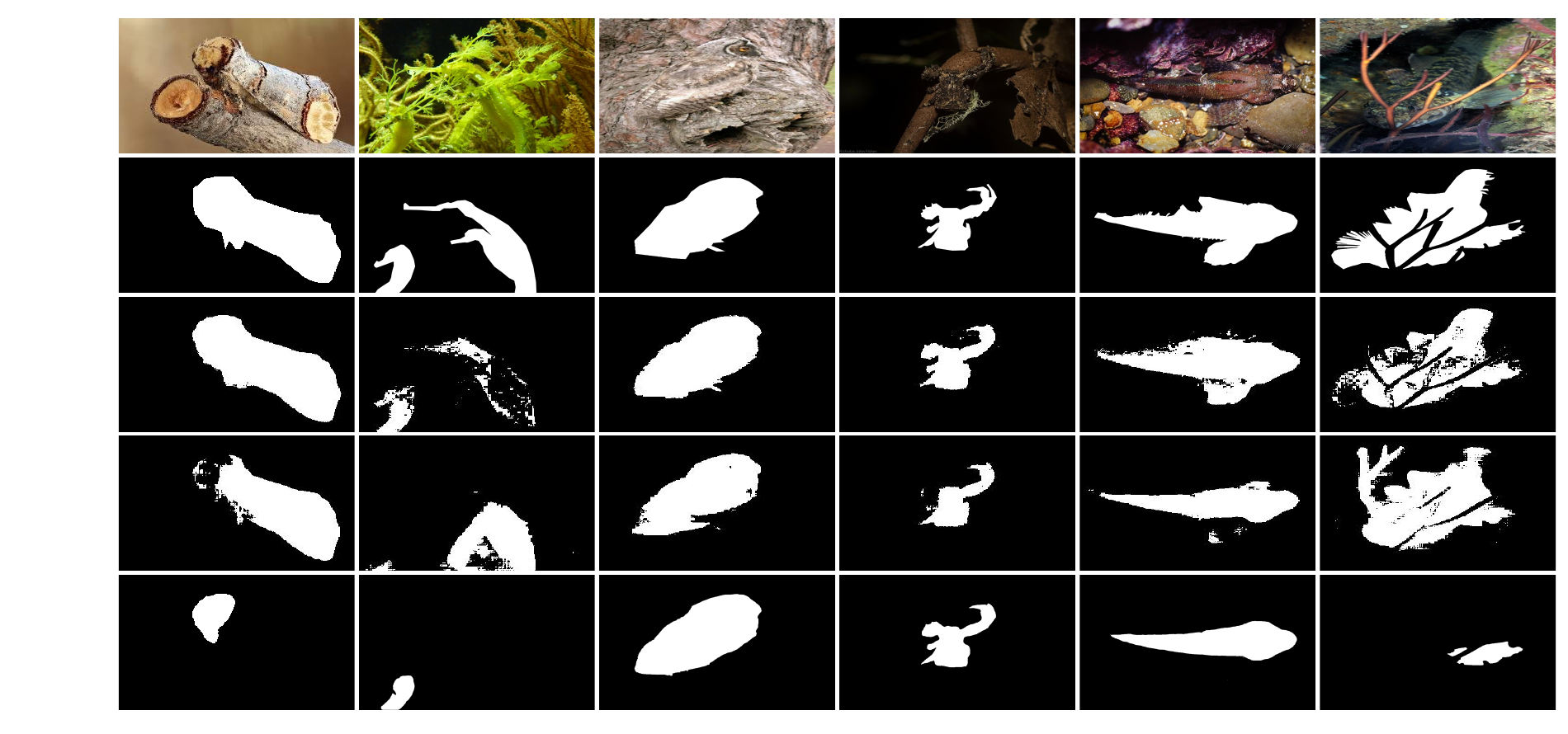} 
    \put(4,39){\footnotesize \rotatebox{90}{\textbf{Image}}} 
    \put(4,31){\footnotesize \rotatebox{90}{\textbf{GT}}} 
    \put(4,21){\footnotesize \rotatebox{90}{\textbf{Ours}}}
    \put(4,10.5){\footnotesize \rotatebox{90}{\textbf{MedSAM}}} 
    \put(4,4){\footnotesize  \rotatebox{90}{\textbf{SAM}}} 
\end{overpic}
\vspace{-5pt}
\caption{\small{\textbf{Comparison of our COMPrompter and other methods, including MedSAM~\cite{ma2023segment} and SAM~\cite{Kirillov_2023_ICCV}, in terms of COD}. Columns 1--3 are for the CAMO dataset, and Columns 4--6 are fromfor the COD10K dataset.}}
\label{fig:quality_CAMO_COD10K}
\end{figure*}

\subsection{Discrete wavelet transform}
\vspace{-5pt}
In image processing, DWT can capture features with various frequencies. Among these, the high-frequency features represent edges and subtle changes in the image. Wang \etal~\cite{wang2022objectformer} extracted the high-frequency features of the an image. Liu \etal~\cite{Liu2023explicit} used a high-frequency component as a prompt to adapt to various downstream tasks.

Inspired by He \etal~\cite{he2023camouflaged}, we apply DWT to the image embedding. DWT focuses on diagonal high-frequency regions in the image. Specifically, it captures rapid changes in signal in the diagonal direction. DWT obtains the diagonal high frequency via the diagonal difference of the original signal. The diagonal high-frequency information (HF) is defined as follows:
\begin{equation} \label{equ:DWT}
HF = x_1 - x_2 - x_3 + x_4,
\end{equation}
where $x_1$ and $x_2$ represent the horizontal and vertical components of the low-frequency signal,respectively, and $x_3$ and $x_4$ represent the horizontal and vertical components of the high-frequency signal, respectively. The high frequency extracted via DWT is added to BBMG. The $OBB$ is connected with $HF$ as per dimension. The adapted pointwise convolution of depthwise separable
convolutions~\cite{Laurent2014Rigid} is then performed. The optimized dense embedding ($ODE$) is generated as follows:
\begin{equation} 
\label{equ:DWT_after}
ODE=DC\big(cat(HF,OBB)\big),
\end{equation}
where $cat (\cdot)$ denotes the join operation by channel dimension, and $DC (\cdot)$ an adapted pointwise convolution of depthwise separable convolutions~\cite{Laurent2014Rigid}.
\begin{table*}[t!]
\centering
\footnotesize
\renewcommand{\arraystretch}{1.2}
\setlength\tabcolsep{1pt}
\begin{tabular}{l*{21}{c}}
\toprule
\multirow{2}{*}{Dataset} & \multirow{2}{*}{Metric} 
& \makecell[c]{UC\\Net} 
& \makecell[c]{SI\\Net} 
& \makecell[c]{Pra\\Net} 
& \makecell[c]{C2F\\Net}       
& \makecell[c]{TI\\Net} 
& \makecell[c]{UG\\TR}
& \makecell[c]{PF\\Net}         
& \makecell[c]{R\\MGL}         
& LSR  
& \makecell[c]{JC\\SOD} 
& \makecell[c]{SIN\\etV2}     
& \makecell[c]{Zoom\\Net} 
& \makecell[c]{Seg\\MaR}        
& \makecell[c]{DG\\Net} 
& \makecell[c]{MSC\\AFNet}     
& SAM  
& \makecell[c]{SAMA\\dapter} 
& \makecell[c]{Med\\SAM}       
& \textbf{Ours*}
& \textbf{Ours}\\
\cmidrule(r){3-22} 
&& \makecell[c]{2020\\ \cite{zhang2020uc}}    & \makecell[c]{2020\\ \cite{9156837}} 
& \makecell[c]{2020\\ \cite{fan2020pranet}}   & \makecell[c]{2021\\ \cite{sun2021context}}
& \makecell[c]{2021\\ \cite{zhu2021inferring}}     & \makecell[c]{2021\\ \cite{yang2021uncertainty}}
& \makecell[c]{2021\\ \cite{mei2021camouflaged}}     
& \makecell[c]{2021\\ \cite{zhai2021mutual}}     & \makecell[c]{2021\\ \cite{lv2021simultaneously}}
& \makecell[c]{2021\\ \cite{li2021uncertainty}}     & \makecell[c]{2022\\ \cite{fan2021concealed}}
& \makecell[c]{2022\\ \cite{pang2022zoom}}     & \makecell[c]{2022\\ \cite{jia2022segment}}
& \makecell[c]{2023\\ \cite{ji2023deep}}   & \makecell[c]{2023\\ \cite{liu2023mscaf}}
& \makecell[c]{2023\\ \cite{Kirillov_2023_ICCV}}     & \makecell[c]{2023\\ \cite{chen2023sam}}
& \makecell[c]{2023\\ \cite{ma2023segment}} & - & -\\
\midrule
\multirow{4}{*}{\centering CAMO} 
& $F_\beta^\omega \uparrow$
& .700 & .644 & .663 & .719 
& .678 & .684 & .695  
& .673 & .696 & .728 & .743 
& .752 & .742 & .769 & \underline{.828} 
& .606 & .765 & .779 & \underline{.819} & \textbf{.858}  \\
& $S_\alpha \uparrow$
& .739 & .745 & .769 & .796
& .781 & .784 & .782 
& .775 & .787 & .800 & .820 
& .820 & .815 & .839 & \underline{.873}
& .684 & .847 & .820 & \underline{.853} & \textbf{.882} \\
& $E_\phi \uparrow$
& .787 & .829 & .837 & .864 
& .848 & .851 & .842  
& .847 & .854 & .873 & .882 
& .892 & .872 & .901 & \underline{.929} 
& .687 & .873 & .904 & \underline{.919} & \textbf{.942} \\
& $M \downarrow$
& .095 & .092 & .094 & .080 
& .087 & .086 & .085  
& .088 & .080 & .073 & .070
& .066 & .071 & .057 & \underline{.046}
& .132 & .070 & .065 & \underline{.054} & \textbf{.044} \\
\hline
\multirow{4}{*}{\centering \makecell[l]{CHAM-\\ELEON} } 
& $F_\beta^\omega \uparrow$
& .836 & .806 & .763 & .828 
& .783 & .794 & .810  
& .813 & .839 & .848 & .816 
& .845 & \underline{.860} & .816 & \textbf{.865} 
& .639 & .824 & .813 & .830 & \underline{.857}  \\
& $S_\alpha \uparrow$
& .880 & .872 & .860 & .888
& .874 & .888 & .882 
& .893 & .893 & .894 & .888 
& \underline{.902} & \underline{.906} & .890 & \textbf{.912}
& .727 & .896 & .868 & .884 & \underline{.906} \\
& $E_\phi \uparrow$
& .930 & .946 & .907 & .935 
& .916 & .940 & .931  
& .923 & .938 & .943 & .942 
& \textbf{.958} & \underline{.954} & .934 & \textbf{.958} 
& .734 & .919 & .936 & .946 & \underline{.955} \\
& $M \downarrow$
& .036 & .034 & .044 & .032 
& .038 & .031 & .033  
& .030 & .033 & .030 & .030 
& \underline{.023} & \underline{.025} & .029 & \textbf{.022}
& .081 & .033 & .036 & .030 & .026 \\
\hline
\multirow{4}{*}{\centering COD10K} 
& $F_\beta^\omega \uparrow$
& .681 & .631 & .629 & .686 
& .635 & .667 & .660  
& .666 & .673 & .684 & .680 
& .729 & .724 & .693 & .775
& .701 & \underline{.801} & .751 & \underline{.779} & \textbf{.821} \\
& $S_\alpha \uparrow$
& .776 & .776 & .789 & .813 
& .793 & .818 & .800  
& .814 & .804 & .809 & .815 
& .838 & .833 & .822 & \underline{.865}
& .783 & \underline{.883} & .841 & .861 & \textbf{.889} \\
& $E_\phi \uparrow$
& .857 & .864 & .861 & .890 
& .861 & .853 & .877  
& .852 & .880 & .884 & .887 
& .911 & .895 & .896 & \underline{.927}
& .798 & .918 & .917 & \underline{.933} & \textbf{.949} \\
& $M \downarrow$
& .042 & .043 & .045 & .036 
& .042 & .035 & .040  
& .035 & .037 & .035 & .037 
& .029 & .033 & .033 & \underline{.024}
& .049 & \underline{.025} & .033 & .026 & \textbf{.023} \\
\hline
\multirow{4}{*}{\centering NC4K}
& $F_\beta^\omega \uparrow$
& .777 & .723 & .724 & .762 
& .734 & .747 & .745  
& .731 & .766 & .771 & .770 
& .784 & .781 & .784 & \underline{.839}
& .696 & -    & .821 & \underline{.840} & \textbf{.876} \\
& $S_\alpha \uparrow$
& .813 & .808 & .822 & .838 
& .829 & .839 & .829 
& .833 & .840 & .842 & .847 
& .853 & .841 & .857 & \underline{.887}
& .767 & -    & .866 & \underline{.880} & \textbf{.907} \\
& $E_\phi \uparrow$
& .872 & .871 & .876 & .897 
& .879 & .874 & .888  
& .867 & .895 & .898 & .903 
& .912 & .905 & .911 & \underline{.935}
& .776 & -    & \underline{.929} & \underline{.935} & \textbf{.955} \\
& $M \downarrow$
& .055 & .058 & .059 & .049 
& .055 & .052 & .053  
& .052 & .048 & .047 & .048 
& .043 & .046 & .042 & \underline{.032}
& .078 & -    & .041 & \underline{.036} & \textbf{.030} \\
\bottomrule
\end{tabular}
\caption{\textbf{Quantitative results on four different datasets: CAMO, CHAMELEON, COD10K, and NC4K.} The scores in \textbf{bold} represent the best results, while the \underline{underlined} scores indicate the second and third best results}. $\uparrow$ indicates that the higher the score the better and $\downarrow$ indicates that the lower the score the better.
\label{tab:tab_COD}
\end{table*}

\begin{table}[t!]
\centering
\footnotesize
\renewcommand{\arraystretch}{1.0}
\setlength\tabcolsep{8pt}
\begin{tabular}{lcccccccccc}
\toprule
Dataset & \multicolumn{2}{c}{Kvasir}& 
\multicolumn{2}{c}{\makecell[c]{CVC-linicDB}} & \multicolumn{2}{c}{\makecell[c]{CVC-ColonDB}} 
& \multicolumn{2}{c}{\makecell[c]{CVC-300}} & \multicolumn{2}{c}{\makecell[c]{ETIS-LaribPolypDB}} \\
\cmidrule(r){2-3} \cmidrule(r){4-5} \cmidrule(r){6-7} \cmidrule(r){8-9} \cmidrule(r){10-11} 
Metric& $mD \uparrow$ & $mI \uparrow$
& $mD \uparrow$ & $mI \uparrow$
& $mD \uparrow$ & $mI \uparrow$
& $mD \uparrow$ & $mI \uparrow$
& $mD \uparrow$ & $mI \uparrow$   \\
\midrule
U-Net~\cite{ronneberger2015u}
& 0.818 & 0.746
& 0.823 & 0.755
& 0.512 & 0.444
& 0.710 & 0.627
& 0.398 & 0.335 \\
UNet++~\cite{zhou2018unet++}
& 0.821 & 0.743
& 0.794 & 0.729
& 0.483 & 0.410
& 0.707 & 0.624
& 0.401 & 0.344 \\
\makecell[l]{ResU-Net++~\cite{jha2019resunet++}} 
& 0.813 & 0.793
& 0.796 & 0.796
& - & -
& - & -
& - & - \\
SFA~\cite{fang2019selective}
& 0.723 & 0.611
& 0.700 & 0.607
& 0.469 & 0.347
& 0.467 & 0.329
& 0.297 & 0.217 \\
PraNet~\cite{fan2020pranet} 
& 0.898 & 0.840
& 0.899 & 0.849
& 0.709 & 0.640
& 0.871 & 0.797
& 0.628 & 0.567 \\
EU-Net~\cite{patel2021enhanced} 
& 0.908 & 0.854
& 0.902 & 0.846
& 0.756 & 0.681
& 0.837 & 0.765
& 0.687 & 0.609 \\
SANet~\cite{wei2021shallow} 
& 0.904 & 0.847
& 0.916 & 0.859
& 0.753 & 0.670
& 0.888 & 0.815
& 0.750 & 0.654 \\
LDNet~\cite{zhang2022lesion} 
& 0.902 & 0.847
& 0.909 & 0.856
& 0.752 & 0.678
& 0.850 & 0.781
& 0.605 & 0.542 \\
FAPNet~\cite{zhou2022feature} 
& 0.902 & 0.849
& 0.925 & 0.877
& 0.731 & 0.658
& 0.893 & 0.826
& 0.717 & 0.643  \\
\midrule
SAM~\cite{Kirillov_2023_ICCV} 
& 0.799 & 0.720
& 0.580 & 0.518
& 0.488 & 0.423
& 0.669 & 0.614
& 0.538 & 0.488  \\
\makecell[l]{Med-SAM~\cite{ma2023segment}} 
& 0.909 & 0.857
& 0.916 & 0.858
& 0.877 & 0.798
& 0.914 & 0.848
& 0.855 & 0.783  \\
\makecell[l]{\textbf{Ours}}
& \textbf{0.935} & \textbf{0.892}
& \textbf{0.931} & \textbf{0.883}
& \textbf{0.917} & \textbf{0.856}
& \textbf{0.938} & \textbf{0.889}
& \textbf{0.910} & \textbf{0.849}  \\
\bottomrule
\end{tabular}
\caption{\small{\textbf{Quantitative results on five different datasets: CVC-ClinicDB, Kvasir, CVC-300, CVC-ColonDB, and ETIS-LaribPolypDB.} The scores in \textbf{bold} are the best ones.} $mD$ represents mean dice similarity coefficient ($mDice$) and $mI$ denotes mean Intersection over Union ($MIoU$). $\uparrow$ indicates that the higher the score the better, and $\downarrow$ indicates that the lower the score the better.}
\label{tab:tab_polyp}
\end{table}
\section{Experiments}\label{sec:experi}
\vspace{-5pt}
\subsection{Datatset}
\vspace{-5pt}
We conducted experiments on four widely recognized datasets, namely CAMO~\cite{le2019anabranch}, CHAMELEON~\cite{skurowski2018animal}, COD10K~\cite{fan2021concealed}, and NC4K~\cite{lv2021simultaneously}, to examine the effect of COMPrompter in the task of COD.
CAMO comprised 1250 images, randomly split into a training dataset of 1000 images and a test dataset of 250 images. CHAMELEON had 76 images for COD. COD10K had 5066 images, of which 3040 were of a training dataset and 2026 of a test dataset. NC4K was fairly large, with 4121 images. It was used as a test dataset for experiments to examine the generalization ability of COMPrompter.
Following Fan \etal~\cite{fan2021concealed}, we adopted 3040 images of COD10K and 1000 images of CAMO as the training dataset. The remaining images of COD10K and CAMO, the entire NC4K dataset and the entire CHAMELEON dataset were used as the test dataset.
In addition, we tested COMPrompter on a more specific application, polyp segmentation, for a more in-depth evaluation. Following Fan \etal~\cite{fan2020pranet}, we used five public benchmarks datasets, ETIS-Larib~\cite{silva2014toward}, CVC-ClinicDB~\cite{bernal2015wm}, CVC-ColonDB~\cite{tajbakhsh2015automated}, CVC300~\cite{vazquez2017benchmark}, and Kvasir-SEG~\cite{jha2020kvasir}.
\subsection{Experimental setup}
\vspace{-5pt}
\textbf{Implementation details.}
COMPrompter was implemented using PyTorch, employing the Adam optimizer with a learning rate of $1e^{-5}$. The model underwent 300 epochs to achieve optimal performance. The process was completed in approximately 4.2 h on an NVIDIA 3080TI GPU with a batch size of 32. We scaled all the input images to $1024 \times 1024$ via bilinear interpolation, scaling them either up or down. In addition, we truncated and normalized the input image data. This ensured the pixel values were in the appropriate range while maintaining the relative distribution relationship of the data.

\begin{figure}[t!]
\centering
\begin{overpic}[width=0.75\linewidth]{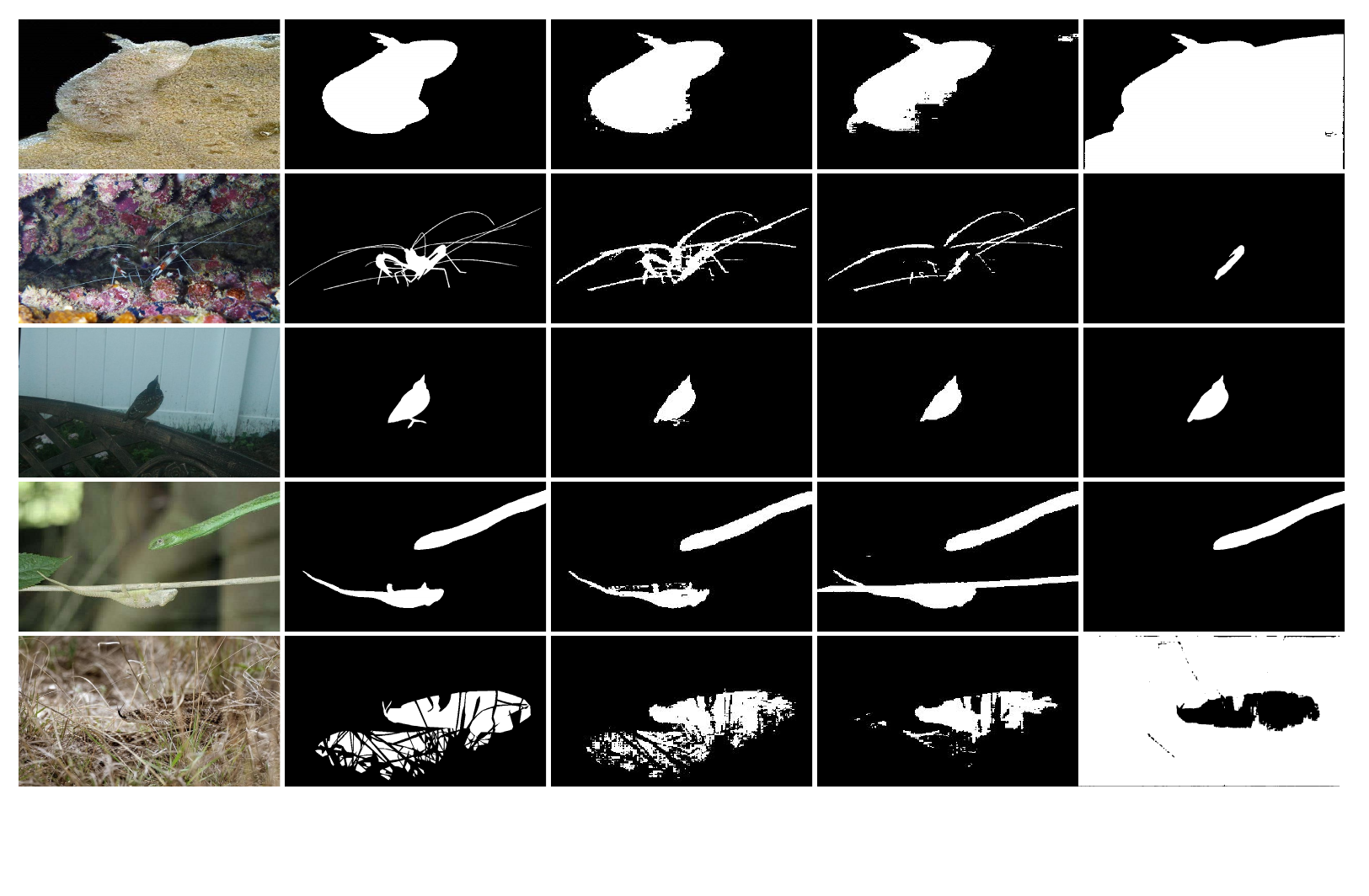} 
    \put(85,4){\footnotesize {\textbf{SAM
}}} 
    \put(63.5,4){\footnotesize {\textbf{MedSAM
}}} 
    \put(41,4){\footnotesize {\textbf{COMPrompter
}}}
    \put(28.5,4){\footnotesize {\textbf{GT}}} 
    \put(8,4){\footnotesize  {\textbf{Image}}} 
\end{overpic}
\vspace{-15pt}
\caption{\small{\textbf{Comparison of our COMPrompter with other methods, including MedSAM~\cite{ma2023segment} and SAM~\cite{Kirillov_2023_ICCV}, in terms of COD}. The selected images are from NC4K and contain various shapes, categories, and camouflage methods.}}
\label{fig:quality_nc4k}
\end{figure}

\begin{figure}[t!]
\centering
\begin{overpic}[width=0.74\linewidth]{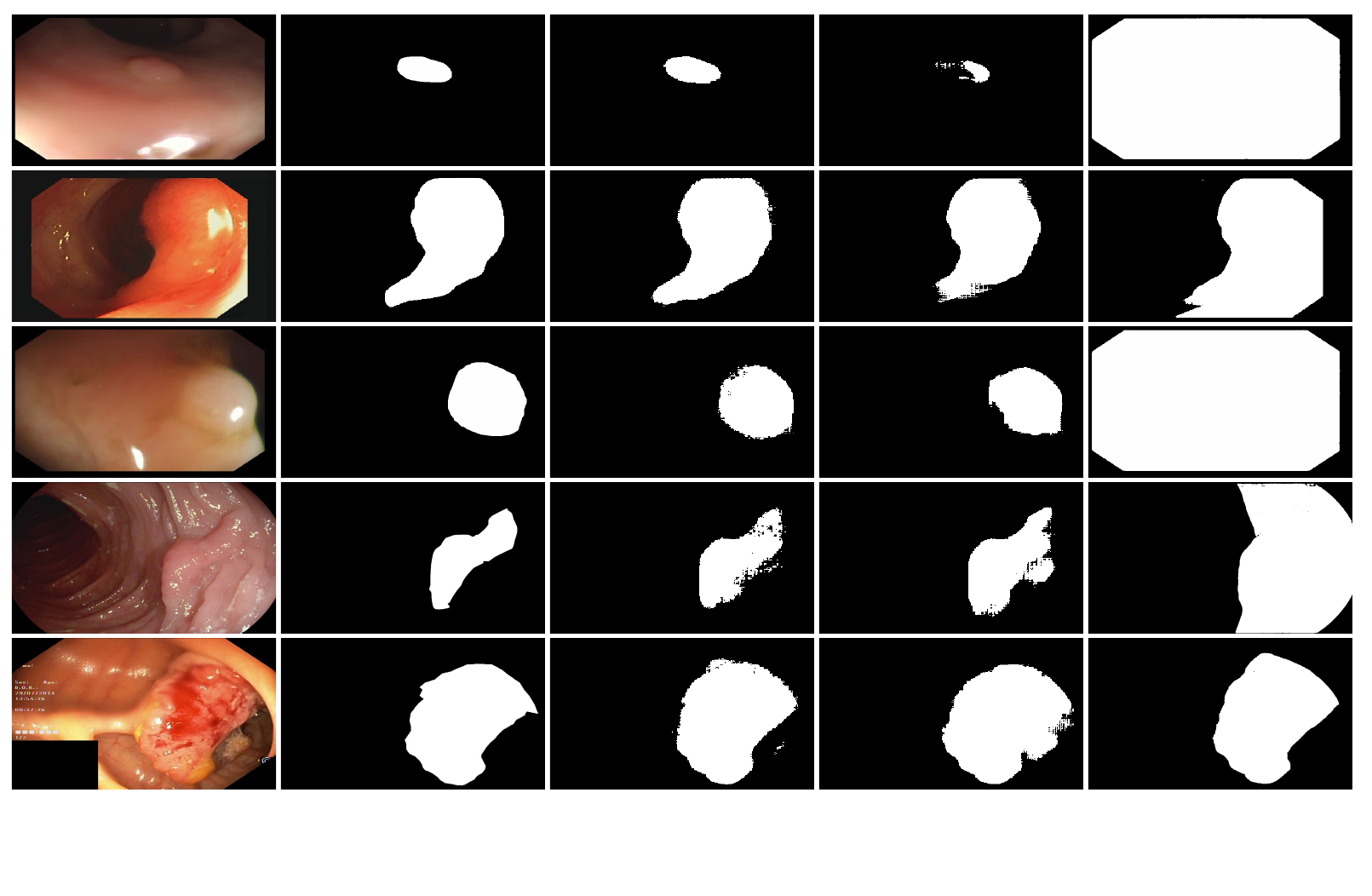} 
    \put(85,4){\footnotesize {\textbf{SAM
}}} 
    \put(63.5,4){\footnotesize {\textbf{MedSAM
}}} 
    \put(40.5,4){\footnotesize {\textbf{COMPrompter
}}}
    \put(28,4){\footnotesize {\textbf{GT}}} 
    \put(7,4){\footnotesize  {\textbf{Image}}} 
\end{overpic}
\vspace{-15pt}
\caption{\small{\textbf{Comparison of COMPrompter with other methods, including \eg, MedSAM~\cite{ma2023segment} and SAM~\cite{Kirillov_2023_ICCV}, in polyp datasets.} We have provided different examples for a comprehensive comparison.}}
\label{fig:quality_polyp}
\end{figure}
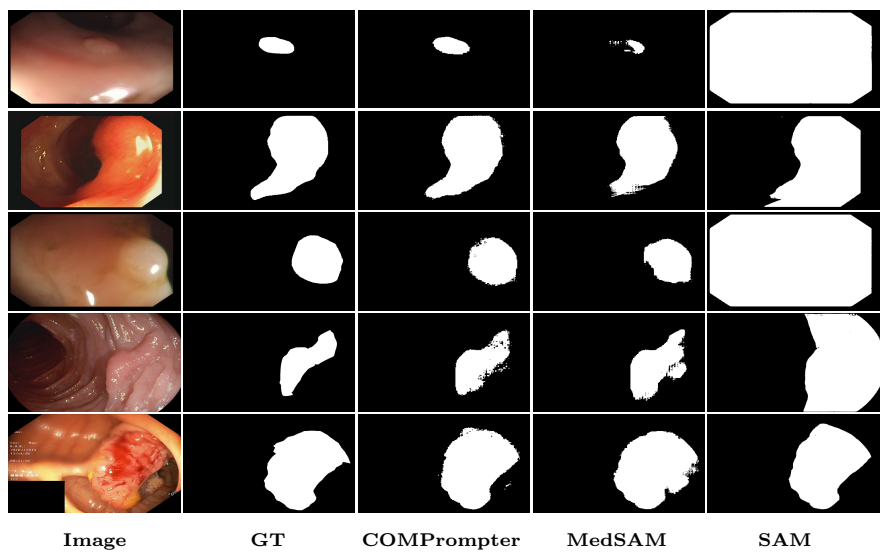

\begin{table}[t]
\centering
\footnotesize
\renewcommand{\arraystretch}{0.9}
\setlength\tabcolsep{15pt}
\begin{tabular}{lcccccccc}
\toprule
\multicolumn{9}{c}{Ablation study} \\
\midrule
\multirow{2}{*}{Model} & \multicolumn{4}{c}{COD10K} & \multicolumn{4}{c}{NC4K} \\
\cmidrule(r){2-5} \cmidrule(r){6-9} & 
$F_\beta^\omega$ & $S_\alpha$ & $E_\phi$ & $M$ &
$F_\beta^\omega$ & $S_\alpha$ & $E_\phi$ & $M$   \\
\midrule
M1
& 0.701 & 0.783 & 0.798 & 0.049
& 0.696 & 0.767 & 0.776 & 0.078 \\
M2
& 0.752 & 0.841  & 0.918 & 0.034
& 0.819 & 0.865  & 0.928 & 0.042 \\
M3
& 0.813 & 0.887  & 0.948 & 0.024
& 0.867 & 0.903  & 0.953 & 0.032 \\
M4
& 0.813 & 0.884  & 0.947 & 0.024
& 0.873 & 0.905  & 0.955 & 0.030 \\
M5
& 0.821 & 0.889 & 0.949 & 0.023
& 0.876 & 0.907 & 0.955 & 0.030 \\
\midrule
\multicolumn{9}{c}{Frequencies comparative experiments} \\
\midrule
\multirow{2}{*}{Setting}  & \multicolumn{4}{c}{COD10K} & \multicolumn{4}{c}{NC4K} \\
\cmidrule(r){2-5} \cmidrule(r){6-9}  
& $F_\beta^\omega$ & $S_\alpha$ & $E_\phi$ & $M$ 
& $F_\beta^\omega$ & $S_\alpha$ & $E_\phi$ & $M$   \\
\midrule
$LL$ 
& 0.814 & 0.885 & 0.946 & 0.024 
& 0.871 & 0.904 & 0.954 & 0.031 \\
$LH$ 
& 0.812 & 0.883 & 0.946 & 0.025 
& 0.869 & 0.902 & 0.953 & 0.032 \\
$HL$  
& 0.818 & 0.887 & 0.949 & 0.024 
& 0.873 & 0.905 & 0.955 & 0.031 \\
$HH$ 
& 0.821 & 0.889 & 0.949 & 0.023 
& 0.876 & 0.907 & 0.955 & 0.030 \\
\bottomrule
\end{tabular}
\caption{\small{\textbf{Ablation study for each module of the proposed COMPrompter on COD datasets and the comparative experiments with different frequencies obtained using the DWT module.} \textbf{SAM (M1)}: this setting is the mode for segmenting all object in SAM. We evaluated the mask with the best segmentation quality in this mode as the final result of segmentation. \textbf{SAM + Box (M2)} : providing box prompt guidance based on M1. \textbf{SAM + Boundary (M3)}: providing boundary prompt guidance on the basis of M1. \textbf{SAM + Box + Boundary (M4)}: providing boundary prompt and box prompt on the basis of M1. \textbf{SAM + Box + Boundary + DWT (M5)}: adding DWT module based on M4.} $LL$ denotes the low-frequency part of image. $HH$ denotes high-frequency part in the diagonal direction. $LH$ denotes the combination of the high-frequency part in the horizontal direction and the low-frequency part in the vertical direction. $HL$ denotes the converse.}
\label{tab:tab_abl}
\end{table}

\textbf{Evaluation metrics.}
We adopted four metrics from COD10K~\cite{fan2021concealed} that are widely used and recognized in the field of COD: structure measure ($S_\alpha$), weighted F-measure ($F_\beta^\omega$), mean enhanced-alignment measure ($E_\phi$), and mean absolute error ($M$). In the polyp segmentation experiment, we selected the mean dice similarity coefficient ($mDice$) and mean Intersection over Union ($mIoU$). The structure measure quantifies the structural similarity between predicted results and actual segmented regions. The weighted F-measure combines precision and recall, and weights them. The enhanced-alignment measure evaluates prediction results by comparing the alignment relationship between the predicted value and the actual value. The mean absolute error is a quantification of the mean absolute error between the predicted value and the true value.

\subsection{Comparisons with cutting-edge methods}
\begin{figure*}[h]
\centerline{\includegraphics[width=0.7\textwidth]{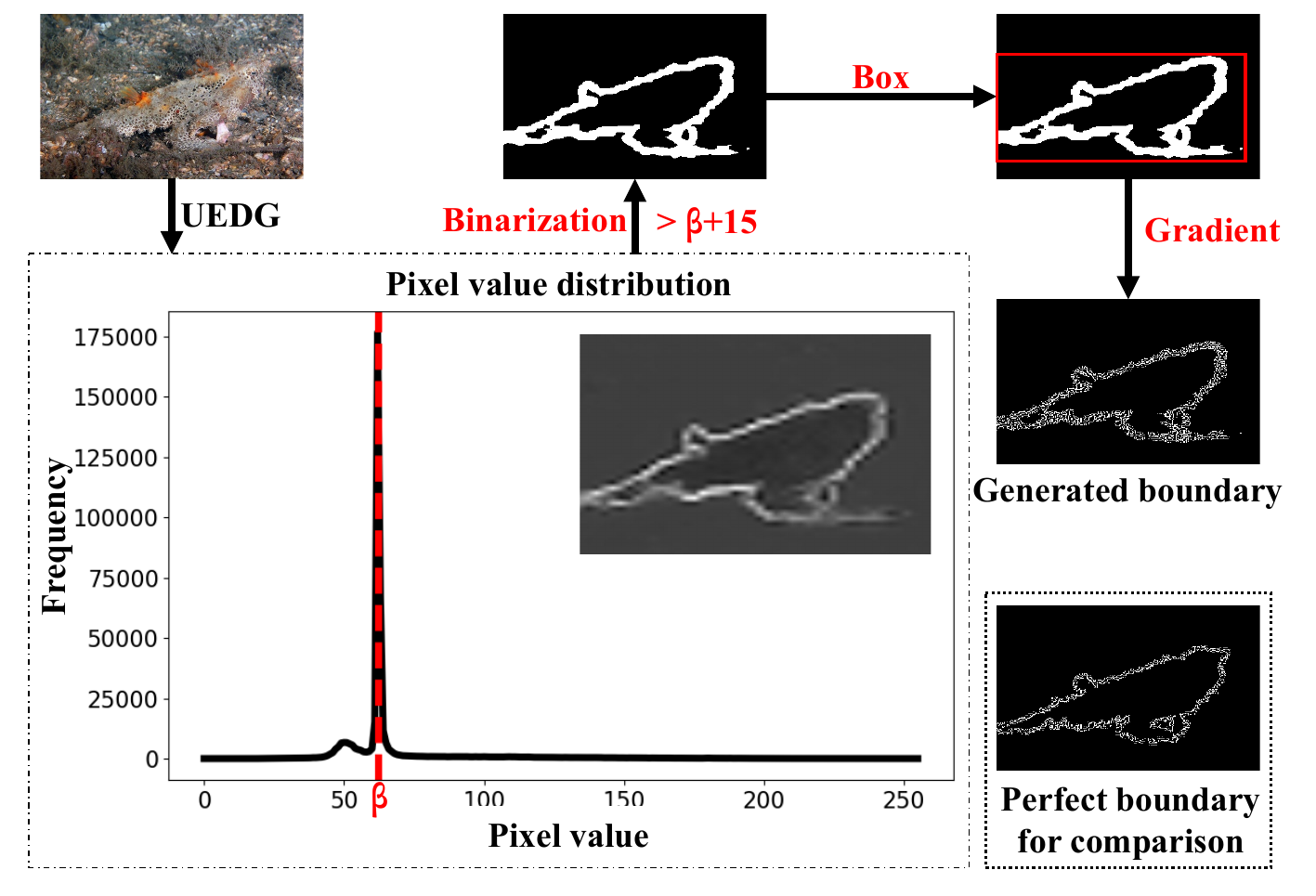}}
\caption{\small{\textbf{Process of the generation of boundary with gradient.} $\beta$ represents the pixel value with the largest proportion, which is unfixed. The number 15 represents the offset value, which is fixed. The key step is in red.}}
\label{fig:fake_boundary}
\end{figure*}

\vspace{-5pt}
We now compare COMPrompter with SAM~\cite{Kirillov_2023_ICCV} and other existing COD algorithms, such as UCNet~\cite{zhang2020uc}, SINet~\cite{9156837}, PraNet~\cite{fan2020pranet},  C2FNet~\cite{sun2021context}, TINet~\cite{zhu2021inferring}, UGTR~\cite{yang2021uncertainty}, PFNet~\cite{mei2021camouflaged},
R-MGL~\cite{zhai2021mutual}, LSR~\cite{lv2021simultaneously}, JCSOD~\cite{li2021uncertainty}, SINetV2~\cite{fan2021concealed}, 
ZoomNet~\cite{pang2022zoom}, SegMaR~\cite{jia2022segment},
DGNet~\cite{ji2023deep}, MSCAF-Net~\cite{liu2023mscaf}, 
SAM-Adapter~\cite{chen2023sam}, and
MedSAM~\cite{ma2023segment}. This qualitative results is shown in Figs.~\ref{fig:quality_CAMO_COD10K}, ~\ref{fig:quality_nc4k}, and ~\ref{fig:quality_polyp}.
To assess the practical usability of COMPrompter, we provided metric data based on the boundaries generated by the pretrained model. These results are shown in the $Our*$ column of Table~\ref{tab:tab_COD}. The procedure of the boundary generation is shown in Fig.~\ref{fig:fake_boundary}. First, we obtained the edges using UEDG~\cite{10183371}. Next, we achieved a clearer mask via binarization with a flexible threshold value. This threshold value comprised the pixel value with the highest percentage (the background pixel value, computed from the histogram) plus an offset value of 15. Thereafter, we set the pixels outside the bounding box to zero as per the box prompt and multiplied the result with the gradient map. Finally, we obtained the generated boundary with a gradient.
Although MedSAM is used in the field of medical image processing, this method can be used to improve SAM for universal applications. Upon applying the algorithm to the COD task, the metric indexes was considerably improved. Hence, we listed MedSAM as one of the comparison algorithms. For MedSAM, we performed retraining and validation on the basis of the official code.
We also compared COMPrompter with the existing methods vis-à-vis polyp segmentation, such as U-Net~\cite{ronneberger2015u}, UNet++~\cite{zhou2018unet++}, ResUNet++~\cite{jha2019resunet++}, SFA~\cite{fang2019selective}, PraNet~\cite{fan2020pranet}, 
EU-Net~\cite{patel2021enhanced}, SANet~\cite{wei2021shallow}, LDNet~\cite{zhang2022lesion}, FAPNet~\cite{zhou2022feature}, SAM~\cite{Kirillov_2023_ICCV}, and MedSAM~\cite{ma2023segment}.
As shown in Table~\ref{tab:tab_COD} and Table~\ref{tab:tab_polyp}, our proposed COMPrompter achieved SOTA performance in the COD and polyp segmentation domains. We later present a detailed qualitative and quantitative analysis of the results in both these domains.

\textbf{Quantitative result.}
COMPrompter introduces detailed prompts and fine-tuning techniques particularly designed for COD tasks. In comparison with SAM, COMPrompter resulted in considerable advancement in the evaluation metrics. Compared with SAM-Adapter, which is another SAM-based COD method, COMPrompter achieved distinction with several advantages. On the COD10K dataset, COMPrompter achieved enhancements, including a $2\%$ rise in $F_\beta^\omega$, $0.6\%$ increase in $S_\alpha$, $3.1\%$ boost in $E_\phi$, and $0.2\%$ improvement in $M$ versus SAM-Adapter. COMPrompter achieved an average improvement of 4.9\%, 1.7\%, 4.5\%, and 1.2\% across $F_\beta^\omega$, $S_\alpha$, $E_\phi$, and $M$, respectively, on three datasets in comparison with SAM-Adapter.
COMPrompter demonstrates its superiority to non-SAM methods as well. On the CAMO dataset, COMPrompter outperformed MSCAFNet with a $3\%$ boost in $F_\beta^\omega$, $0.9\%$ enhancements in $S_\alpha$, $1.3\%$ progress in $E_\phi$, and $0.2\%$ increase in $M$. From a holistic dataset viewpoint, COMPrompter exhibited an average enhancement of 2.6\%, 1.2\%, 1.3\%, and 0.03\% across $F_\beta^\omega$, $S_\alpha$, $E_\phi$, and $M$, respectively, compared with MSCAFNet on the four evaluated datasets. When compared with MSCAFNet on the CHAMELEON dataset, COMPrompter demonstrated certain shortcomings. Overall, these results highlighted the effectiveness of COMPrompter in COD tasks. The experimental results of polyp segmentation are presented in Table~\ref{tab:tab_polyp}. Compared with MedSAM, COMPrompter had an average gain of $3.2\%$ in $mDice$ and $4.5\%$ in $mIoU$. For COMPrompter with the generated boundary, the accuracy ranked within the top three.

\textbf{Qualitative results.}
With a view to more intuitively showing the segmentation effect of our proposed COMPrompter on COD datasets and polyp datasets, we compared the original image and GT with the predictions generated by COMPrompter, MedSAM, and SAM. This comparison is shown in Figs.~\ref{fig:quality_CAMO_COD10K}, ~\ref{fig:quality_nc4k}, and ~\ref{fig:quality_polyp}.
Figures~\ref{fig:quality_CAMO_COD10K} and ~\ref{fig:quality_nc4k} show the learning ability and generalization ability of COMPrompter, respectively. Because there is a lack of certain semantic information when SAM is directly applied to COD, only a part of the target was segmented. This is illustrated in the first and last columns in Fig.~\ref{fig:quality_CAMO_COD10K}. Because of providing a strong prior, the segmentation effect of MedSAM was observed to be greatly improved. In view of the fact that the bounding box only provides an approximate location, there still exists a certain level of semantic ambiguity. This resulted in occasional segmentation errors, as shown in the second image of Fig.~\ref{fig:quality_CAMO_COD10K} for MedSAM. In addition, the processing of details of the edges and occlusions of the target was not particularly ideal. In particular, in the fifth line in Fig.~\ref{fig:quality_nc4k}, we can see that MedSAM is affected by the weed and does not segment out the target hidden behind it. However, our proposed COMPrompter can segment the entire object and consider the details. One can see that the foot of the bird in the third column in Fig.~\ref{fig:quality_CAMO_COD10K} is segmented. In terms of foreground occlusion, COMPrompter can clearly distinguish between target and occlusion, and finely segment them out, as shown in the sixth column of Fig.~\ref{fig:quality_CAMO_COD10K} and the fifth line of Fig.~\ref{fig:quality_nc4k}.

\subsection{Ablation study}
\label{sec:ablation}
\vspace{-8pt}
We confirmed the effectiveness of the box prompt, boundary prompt, and DWT modules via ablation experiments. 
We added the boundary prompt, box prompt, and DWT modules to SAM in turn, using the same experimental setup as for training. In particular, we designed five models to confirm the effectiveness of the three modules. The results on COD10K and NC4K are presented in Table~\ref{tab:tab_abl}. In general, each module played a positive role in boosting the experimental results.  Finally, our proposed COMPrompter achieved SOTA performance.

As shown in Fig.~\ref{fig:quality_polyp}, the direct application of SAM to polyp segmentation resulted in a large number of incorrectly segmented regions. Although MedSAM can could roughly segment the polyp, the edge was not clear. COMPrompter added boundary prompt on the basis of box prompt and more clearly segmented polyps.

\textbf{Efficiency analysis.}
To comprehensively describe our model, we compared its input size, parameters and inference speed with those of the COD-related and polyp-related models. Table~\ref{tab:complexity} shows that the SAM-based model has larger parameters and a longer inference time. Compared with SAM, COMPrompter greatly decreased the number of parameters and increased the inference speed by four times. Most importantly, the segmentation ability of COMPrompter greatly exceeded that of the existing models, irrespective of whether they were based on SAM or not. The inference times in the table were obtained via testing in one NVIDIA RTX 3080TI GPU. Except for SAM, one can refer to the model performance metrics on the official website.

\begin{table}
\centering
\footnotesize
\renewcommand{\arraystretch}{0.9}
\setlength\tabcolsep{6pt}
\begin{tabular}{lccclccc}
\toprule
\makecell[l]{Methods} & input size & \makecell[c]{Param (M)} & \makecell[c]{Speed (fps)} & \makecell[l]{Methods (COD)} & input size & \makecell[c]{Param (M)} & \makecell[c]{Speed (fps)} \\
\midrule
ResUNet++ & 256 $ \times $ 256 & 4.06 & 1 & 
PFNet & 416 $ \times $ 416 & 46.50 & 46     \\  
PraNet & 352 $\times$ 352 & 32.55 & 42  &
R-MGL & 473 $ \times $ 473 & 67.64 & 14   \\
EU-Net & 256 $\times$ 256 & 31.43 & 11  & 
LSR & 352 $\times$ 352 & 57.90 & 83   \\
SANet & 352 $\times$ 352 & 23.90 & 67  & 
JCSOD & 352 $\times$ 352 & 121.63 & 53    \\
LDNet & 256 $\times$ 256 & 33.38 & 20  & 
SINetV2 & 352 $ \times$ 352 & 26.98 & 50   \\
FAPNet & 352 $ \times $ 352 & 29.52 & 38 & DGNet & 352 $ \times $ 352 & 21.02 & 40  \\
\cmidrule(r){0-3}
PraNet & 352 $ \times $ 352 & 32.55 & 42  & 
SAM & 1024 $\times$ 1024 & 615 & 2       \\
C2FNet & 352 $\times$ 352 & 28.41 & 43 &
MedSAM & 1024 $\times$ 1024 & 93.73 & 8  \\
\midrule
\textbf{Ours} & 1024 $\times$ 1024 & 94.86 & 8 & 
\textbf{Ours} & 1024 $\times$ 1024 & 94.86
& 8 \\
\bottomrule
\end{tabular}
\caption{\small{\textbf{Comparison of network complexity.}}}
\label{tab:complexity}
\end{table}

\begin{table}[t]
\centering
\footnotesize
\renewcommand{\arraystretch}{0.9}
\setlength\tabcolsep{10pt}
\begin{tabular}{lcccccccccc}
\toprule
\multirow{2}{*}{Setting} & \multirow{2}{*}{Dilate1} & \multirow{2}{*}{Dilate2} & \multicolumn{4}{c}{COD10K} & \multicolumn{4}{c}{NC4K} \\
\cmidrule(r){4-7} \cmidrule(r){8-11} 
&  & & 
$F_\beta^\omega$ & $S_\alpha$ & $E_\phi$ & $M$ &
$F_\beta^\omega$ & $S_\alpha$ & $E_\phi$ & $M$   \\
\midrule
D1 & 3 $ \times $ 3 & 3 $ \times $ 3 
& 0.824 & 0.891 & 0.951 & 0.023
& 0.875 & 0.906 & 0.953 & 0.030 \\
D2 (office) & 3 $\times$ 3 & 5 $\times$ 5
& 0.821 & 0.889 & 0.949 & 0.023
& 0.876 & 0.907 & 0.955 & 0.030 \\
D3 & 5 $ \times $ 5 & 5 $ \times $ 5
& 0.820 & 0.888  & 0.949 & 0.023
& 0.875 & 0.907  & 0.954 & 0.030 \\
D4 & 5 $\times$ 5 & 7 $\times$ 7
& 0.817 & 0.886 & 0.947 & 0.024
& 0.874 & 0.906 & 0.954 & 0.030 \\
D5  & 7 $\times$ 7 & 7 $\times$ 7
& 0.814 & 0.884  & 0.948 & 0.024
& 0.873 & 0.905  & 0.953 & 0.030 \\
\bottomrule
\end{tabular}
\caption{\small{\textbf{Ablation study results for the dilate parameters of EGEM on COD10K and NC4K.} Among the settings, the combinations of $1 \times 1$ and $1 \times 1$, and $1 \times 1$ and $3 \times 3$ have not been adopted because employing a kernel size of 1 is insufficient to capture the respective boundaries.}}
\label{tab:tab_EGEM}
\end{table}
\begin{figure}[t!]
\centering
\begin{overpic}[width=0.9\linewidth]{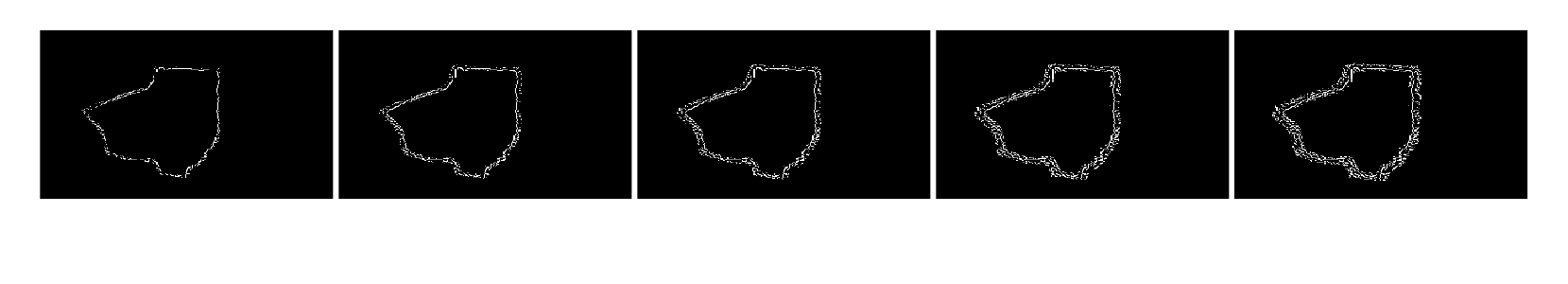} 
    \put(87,3){\footnotesize {\textbf{D5
}}} 
    \put(67.5,3){\footnotesize {\textbf{D4
}}} 
    \put(48,3){\footnotesize {\textbf{D3
}}}
    \put(29.5,3){\footnotesize {\textbf{D2}}} 
    \put(11,3){\footnotesize  {\textbf{D1}}} 
\end{overpic}
\vspace{-16pt}
\caption{\small{\textbf{Comparison of boundary containing gradients obtained for various dilate parameter settings.} One can see  Table~\ref{tab:tab_EGEM} for the dilate parameter settings for D1 -- D5.}}
\label{fig:Abla_EGEM}
\end{figure}
\begin{figure}[t!]
\centering
\begin{overpic}[width=0.7\linewidth]{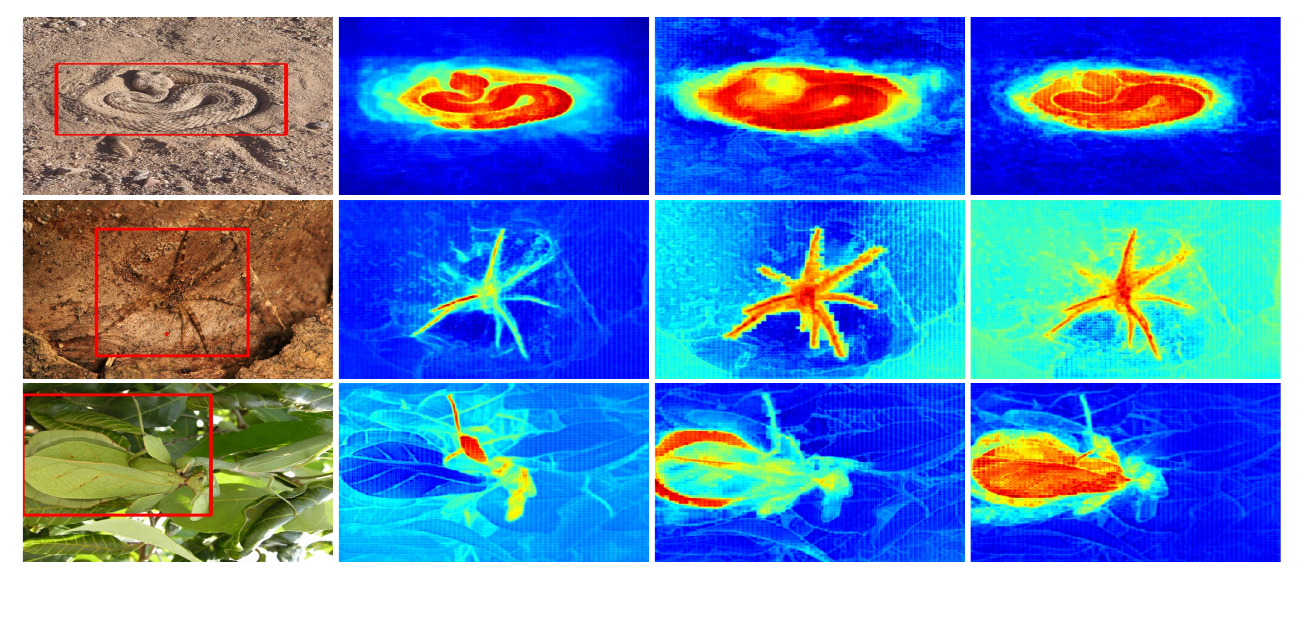} 
    \put(76.5,3){\footnotesize {\textbf{COMPrompter
}}} 
    \put(59.5,3){\footnotesize {\textbf{M3
}}}
    \put(36,3){\footnotesize {\textbf{M2}}} 
    \put(10,3){\footnotesize  {\textbf{Image}}} 
\end{overpic}
\vspace{-10pt}
\caption{\small{\textbf{Feature visualizations for the box prompt condition, the boundary prompt condition, and the full COMPrompter.} Settings for SAM + Box (M2) are used within the box group, while settings for SAM + Boundary (M3) are employed within the boundary group. One can zoom in to see more details.}}
\label{fig:Abla_feature}
\end{figure}
\begin{figure*}[t!]
\centering
\begin{overpic}[width=\linewidth]{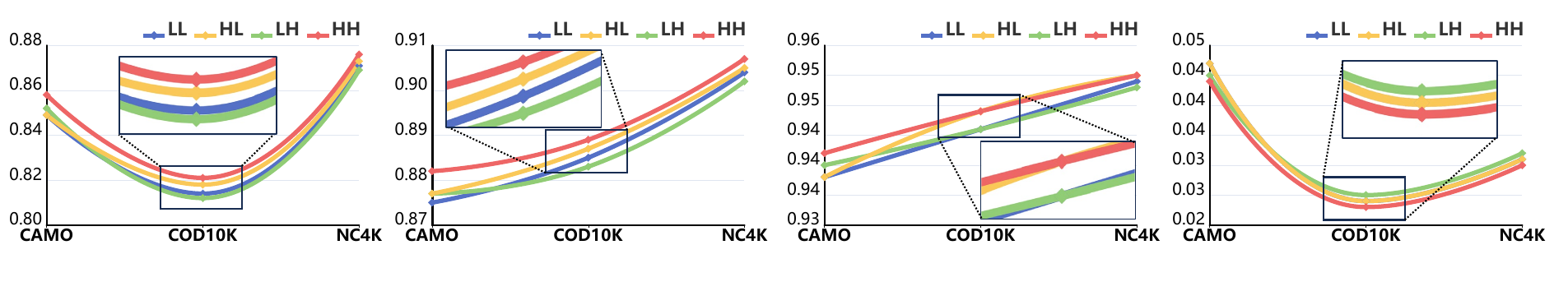} 
    \put(76.5,1){\footnotesize {\textbf{(d) Mean absolute error}}} 
    \put(49,0.5){\footnotesize {\textbf{\makecell[c]{(c) Mean enhanced-alignment \\measure}}}}
    \put(28,1){\footnotesize {\textbf{(b) Structure measure}}} 
    \put(3,1){\footnotesize  {\textbf{(a) Weighted F-measure}}} 
\end{overpic}
\vspace{-13pt}
\caption{\small{\textbf{Line graph with various frequencies obtained using the DWT module}. (a), (b), (c), and (d) represent the performance comparison of $LL$, $LH$, $HL$, and $HH$. This comparison is in terms of the following four evaluation metrics: weighted F-measure, structure measure, mean enhanced-alignment measure, and mean absolute error.}}
\label{fig:DWT_HH}
\end{figure*}
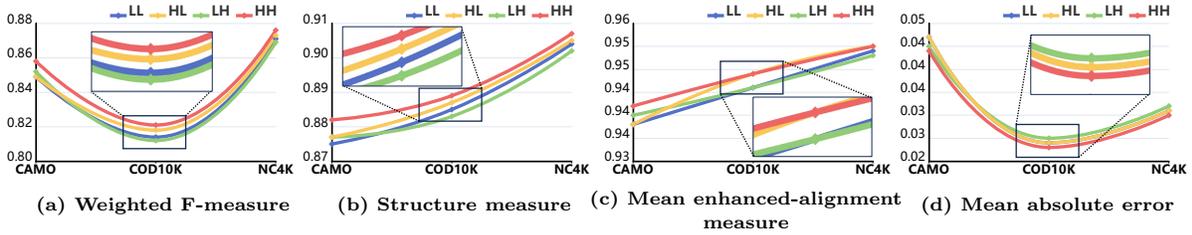
\textbf{Effectiveness of box prompt.}
The effectiveness of the box prompt was confirmed by comparing two pairs of models: from M1 to M2 and from M3 to M4. From M1 to M2, we can see that the box prompt greatly enhanced the model performance. The enhancement achieved on COD10K was $5.1\%$ in $F_\beta^\omega$, while on NC4K, the enhancement was $12.3\%$ in $F_\beta^\omega$. From M3 to M4, other than the guidance already with the boundary prompt, we also saw a performance increase of $0.6\%$ in $F_\beta^\omega$ owing to the introduction of the box prompt in NC4K.

\textbf{Effectiveness of boundary prompt.}
The effectiveness of the boundary prompt was confirmed by comparing two pairs of models: from M1 to M3 and from M2 to M4. Looking at the four metrics, the average augmentation of boundary prompt on two datasets was $14.2\%$ ($F_\beta^\omega$), $12.0\%$ ($S_\alpha$), $16.4\%$ ($E_\phi$), and $3.6\%$ ($M$) from M1 to M3. Compared with M2, M4 achieved an average increase of $5.8\%$ ($F_\beta^\omega$), $4.2\%$ ($S_\alpha$), $2.8\%$ ($E_\phi$), and $1.1\%$ ($M$). These enhancements directly represent the strong effectiveness of the boundary prompt.

In the boundary prompt, EGEM obtains a gradient-containing boundary via appropriate dilation operations. The rationale behind the setting of the dilation parameters is presented in Table~\ref{tab:tab_EGEM}, which shows a general trend of gradual decrease from D1 to D5. However, considering potential biases in boundary acquisition during inference, the dilation parameter setting of D1 seemed overly precise (see Fig.~\ref{fig:Abla_EGEM}). Therefore, we adopted the parameters from D2.

To vividly illustrate the advantages of the multiprompt strategy, we showed the feature maps in the mask decoder of M2, M3, and COMPrompter, as depicted in Fig.~\ref{fig:Abla_feature}. Solely relying on M2 (SAM + Box prompt) yielded feature maps that roughly captured the target but suffered from edge blurriness (first line). In addition, there was insufficient attention to finer details (second line) and incomplete focus (third line). Feature maps generated by M3 (SAM + Boundary) exhibited higher edge attention (second line) but come with a broader activation range (first line). Meanwhile, COMPrompter with the multiprompt strategy demonstrates superior edge attention and appropriate activation ranges and even achieved complementary activation ranges in some instances(third line).

\textbf{Effectiveness of DWT.}
The effectiveness of DWT was confirmed by comparing a pair of models: M4 and M5. Although the improvement brought by DWT is not as notable as that brought by boundary and box prompts, it is still observable. In addition, we conducted comparative experiments on which part of the frequencies in DWT were adopted. The experimental results are presented in Table~\ref{tab:tab_abl}. We have presented the results using a line chart, as shown in Fig.~\ref{fig:DWT_HH}, which shows that $HH$ obtained the best score across all four metrics. Coincidentally, the curves of $LL$ and $HH$ in (d) exactly coincided.

\textbf{Effectiveness of offset value.}
An offset value was used for the binarization of the generated boundary. The effectiveness of the fixed offset value of 15 was demonstrated via a comparison of paired groups. For ease of comparison, we separately calculated the average positive and average negative metrics. As presented in Table~\ref{tab:tab_abl_pix}, the set with an offset of 15 achieved the highest accuracy, with performance decreasing on either side. Therefore, we selected 15 as the optimal offset value.
\begin{table}[h]
\centering
\footnotesize
\renewcommand{\arraystretch}{0.9}
\setlength\tabcolsep{3.5pt}
\begin{tabular}{lcccccccccccccccccc}
\toprule
\multirow{2}{*}{Offset} & \multicolumn{4}{c}{CAMO} &\multicolumn{4}{c}{CHAMELEON} & \multicolumn{4}{c}{COD10K} & \multicolumn{4}{c}{NC4K} &  \multicolumn{2}{c}{Average}\\
\cmidrule(r){2-5} \cmidrule(r){6-9} \cmidrule(r){10-13}  \cmidrule(r){14-17} \cmidrule(r){18-19}&
$F_\beta^\omega \uparrow$ & $S_\alpha \uparrow$ & $E_\phi \uparrow$& $M \downarrow$ &
$F_\beta^\omega \uparrow$ & $S_\alpha \uparrow$ & $E_\phi \uparrow$& $M \downarrow$ &
$F_\beta^\omega \uparrow$ & $S_\alpha \uparrow$ & $E_\phi \uparrow$& $M \downarrow$ &
$F_\beta^\omega \uparrow$ & $S_\alpha \uparrow$ & $E_\phi \uparrow$& $M \downarrow$ & $Up$ & $Down$ \\
\midrule
+5
& .811 & .847 & .918 & .057
& .814 & .875 & .936 & .034
& .770 & .858 & .931 & .028
& .835 & .879 & .935 & .037 & .8674 & .0390\\
+10
& .816 & .850 & .919 & .054
& .824 & .880 & .942 & .031
& .776 & .860 & .932 & .027
& .840 & .881 & .936 & .037 & .8713 & .0373\\
+15 
& .819 & .853 & .919 & .054
& .830 & .884 & .946 & .030
& .779 & .861 & .933 & .026
& .840 & .880 & .935 & .036 & \underline{.8733} & \underline{.0365}\\
+20
& .817 & .849 & .917 & .055
& .831 & .882 & .945 & .031
& .782 & .861 & .934 & .026
& .840 & .879 & .934 & .037 & .8726 & .0373\\
+25
& .816 & .850 & .918 & .055
& .827 & .880 & .941 & .033
& .783 & .861 & .934 & .026
& .841 & .880 & .934 & .037 & .8721 & .0378\\
\bottomrule
\end{tabular}
\caption{\small{\textbf{Ablation study results for offset value of the gradient-containing generated boundary.}} $Up$ denotes positive metric, and $Down$ negative metric.}
\label{tab:tab_abl_pix}
\end{table}

\section{Conclusion}
\label{sec:Conclusion}
\vspace{-5pt}
We proposed COMPrompter, a novel network designed to advance the development of SAM in COD. It use a multiprompt strategy incorporating a box prompt and boundary prompt for accurate priors. The SOTA performance of COMPrompter was demonstrated on multiple datasets. In addition, we highlighted the challenges suffered by COMPrompter in accurately and completely segmenting multiple objects within a single box. A single box can emphasize nontarget areas between multiple targets, resulting in segmentation errors. A possible solution is to use a box prompt with multiple subboxes to accurately cover all targets. We noted that the potential boundary prompt, might provide a novel perspective for COD and related fields. We expect that COMPrompter can contribute to the advancement of SAM application to COD.
\Acknowledgements{This work was supported in part by the National Natural Science Foundation of China [grant no. U2033210, 62101387, and 62475241], and in part by the Zhejiang Provincial Natural Science Foundation [grant no. LDT23F02024F02].}

\end{document}